\documentclass[letterpaper, 10 pt, conference]{ieeeconf}  

\IEEEoverridecommandlockouts                              

\usepackage{graphicx} 
\usepackage{amsmath}
\usepackage{amsfonts}
\usepackage{mathtools}
\usepackage{amssymb}
\usepackage{algorithm}
\usepackage[noend]{algorithmic}
\usepackage{subfig}
\usepackage{soul}
\usepackage{color}
\usepackage{comment}

\newtheorem{theorem}{\textbf{Theorem}}[section]
 
\newtheorem{problem}[theorem]{\textbf{Problem}}

\newtheorem{definition}[theorem]{\textbf{Definition}}

\newcommand{\norm}[1]{|\!|#1|\!|}

\newcommand{\R}{\mathbb{R}}

\newcommand{\N}{\mathbb{N}}

\newcommand{\X}{\mathbb{X}}

\newcommand{\p}{\mathcal{P}}
\newcommand{\NN}{{\mathcal{N}\mathcal{N}}}
\newcommand{\U}{\mathcal{U}}
\newcommand{\W}{\mathcal{W}}

\newcommand{\Post}{\mathrm{Post}}
\newcommand{\Next}{\mathrm{Next}}

\newcommand{\Int}{\mathrm{Int}}
\newcommand{\safe}{\mathrm{safe}}
\newcommand{\goal}{\mathrm{goal}}

\newcommand{\abs}{\mathrm{abs}}
\newcommand{\ct}{\mathrm{ct}}
\newcommand{\pr}{\mathrm{Pr}}

\renewcommand{\baselinestretch}{0.97}

\title{\LARGE \bf
Provably Safe Model-Based Meta Reinforcement Learning: \\An Abstraction-Based Approach
}

\author{Xiaowu Sun$^{1}$, Wael Fatnassi$^{1}$, Ulices Santa Cruz$^{1}$, and Yasser Shoukry$^{1}$
\thanks{
This work was partially sponsored by the NSF awards \#CNS-2002405 and \#CNS-2013824.
}
\thanks{
$^{1}$ Xiaowu Sun, Wael Fatnassi, Ulices Santa Cruz, and Yasser Shoukry are with Department of Electrical Engineering and Computer Science, University of California, Irvine {\tt\small \{xiaowus,wfatnass,usantacr,yshoukry\}@uci.edu}. 
}%
}

\begin{document}

\maketitle
\thispagestyle{empty}
\pagestyle{empty}

\begin{abstract} 
While conventional reinforcement learning focuses on designing agents that can perform one task, meta-learning aims, instead, to solve the problem of designing agents that can generalize to different tasks (e.g., environments, obstacles, and goals) that were not considered during the design or the training of these agents. In this spirit, in this paper, we consider the problem of training a provably safe Neural Network (NN) controller for uncertain nonlinear dynamical systems that can generalize to new tasks that were not present in the training data while preserving strong safety guarantees. Our approach is to learn a set of NN controllers during the training phase. When the task becomes available at runtime, our framework will carefully select a subset of these NN controllers and compose them to form the final NN controller. Critical to our approach is the ability to compute a finite-state abstraction of the nonlinear dynamical system. This abstract model captures the behavior of the closed-loop system under all possible NN weights, and is used to train the NNs and compose them when the task becomes available. We provide theoretical guarantees that govern the correctness of the resulting NN. We evaluated our approach on the problem of controlling a wheeled robot in cluttered environments that were not present in the training data. 
\end{abstract}

\section{INTRODUCTION}
Meta Reinforcement Learning (meta-RL) refers to algorithms that can leverage experience from previous learning experience to learn how to adapt to new tasks quickly. In other words, while contemporary reinforcement learning focuses on designing agents that can perform one task, meta-RL aims to solve the problem of designing agents that can generalize to different tasks that were not considered during the design or the training of these agents. Given data often covering a distribution of related tasks (e.g., changes in the environments, goals, and the dynamics), meta-RL aims to combine all such experience and use it to design agents that can quickly adapt to unforeseen tasks. To achieve such aim, and without loss of generality, meta-training can be seen as a bi-level optimization problem where one optimization contains another optimization as a constraint~\cite{sinha2017review,franceschi2018bilevel}. The inner optimization corresponds to the classical training of a policy to achieve a particular task while the outer optimization focuses instead on optimizing the meta-representation that generalizes to different tasks~\cite{hospedales2020meta,finn2017model,fernando2018meta,fallah2020provably,liu2019taming}. For a review on the current achievements in the field of Meta-RL, we refer the reader to this survey~\cite{hospedales2020meta}.



While the current successes of meta-RL are undeniable, significant drawbacks of meta-RL in its current form are (i) the \emph{lack of formal guarantees on its ability to generalize to unforeseen tasks} and (ii) the \emph{lack of formal guarantees with regards to its safety}.

In this paper, we confine our attention to reach-avoid tasks (i.e., a robot that needs to reach a goal without hitting obstacles) and propose a framework for meta-RL that can generalize to tasks  (e.g., different environments, obstacles, and goals) that were not present in the training data. The proposed framework results into NN controllers that are \emph{provably safe} with regards to \emph{any} reach-avoid task, which could be unseen during the design of these neural networks. 

Recently, the authors proposed a framework for \emph{provably-correct} training of neural networks~\cite{sun2021safeRL}. In that framework, given an error-free nonlinear dynamical system, a finite-state abstract model that captures the closed-loop behavior under all possible neural network controllers is computed. Using this finite-state abstract model, this framework identifies the subset of NN weights guaranteed to satisfy the safety requirements (i.e., avoiding obstacles). During training, the learning algorithm is augmented with a NN weight projection operator that enforces the resulting NN to be provably safe. To account for the liveness properties  (i.e., reaching the goal), the proposed framework uses the finite-state abstract model to identify candidate NN weights that may satisfy the liveness properties. Using such candidate NN weights, the proposed framework biases the NN training to achieve the liveness specification.


While the previous results reported in~\cite{sun2021safeRL} focused on the case when the task (environment, obstacles, and goal) is known during the training of the NN controller, we extend these results in this paper to account for the case when the task is \emph{unknown} during training. In particular, instead of training one neural network, we train a set of neural networks. To fulfill a set of infinitely many tasks using a finite set of neural network controllers, our approach is to restrict each neural network to some local behavior, yet the composition of these neural networks captures all possible behaviors. Moreover, and unlike the results reported in~\cite{sun2021safeRL}, we consider in this paper the case when the nonlinear dynamical system is only partially known. We evaluated our approach on the problem of steering a wheeled robot and we show that our framework is capable of generalizing to tasks that were not present in the training of the NN controller while guaranteeing the safety of the robot.

\section{PROBLEM FORMULATION}  
\label{sec:formulation}
\subsection{Notation}
Let $\norm{x}$ be the Euclidean norm of the vector $x \in \R^n$, $\norm{A}$ be the induced 2-norm of the matrix $A \in \R^{m \times n}$, and $\norm{A}_{\max} = \underset{i,j}{\max}|A_{ij}|$ be the max norm of the matrix $A$. Given two vectors $x_1 \in \R^{n_1}$ and $x_2 \in \R^{n_2}$, we denote by $(x_1, x_2) \in \R^{n_1 + n_2}$ the column vector $[x_1^\top, x_2^\top]^\top$.  We use $\oplus$ to denote the Minkowski sum, and $\Int(\mathcal{S})$ to denote the interior of the set $\mathcal{S}$. Any Borel space $\mathcal{X}$ is assumed to be endowed with a Borel $\sigma$-algebra, which is denoted by $\mathcal{B}(\mathcal{X})$. We use $\mathbf{1}_\mathcal{S}$ to denote the indicator function of a set $\mathcal{S}$.

\subsection{Dynamical Model and Neural Network Controller}
\label{subsec:model}
We consider discrete-time nonlinear dynamical systems of the form:
\begin{equation}
    \label{eq:dyn}
    x^{(k+1)} = f(x^{(k)}, u^{(k)}) + g(x^{(k)}, u^{(k)}),
\end{equation}
where $x^{(k)} \in \mathcal{X} \subset \R^n$ is the state and $u^{(k)} \in \U$ is the control input at time step $k \in \N$. The dynamical model consists of two parts: the priori known nominal model $f$, and the unknown model-error $g$, which is deterministic 
and captures unmodeled dynamics. Though the model-error $g$ is unknown, we assume it is bounded by a compact set $\mathcal{D} \subset \R^n$, i.e., $g(x, u) \in \mathcal{D}$ for all $x \in \mathcal{X}$ and $u \in \U$. We also assume both functions $f$ and $g$ are locally Lipschitz continuous. As a well-studied technique to learn unknown functions from data, we assume the model-error $g$ can be learned using Gaussian Process (GP) regression~\cite{GP}. We use $\mathcal{G}\mathcal{P}(\mu_g, \sigma^2_g)$ to denote a GP regression model with the posterior mean and variance functions be $\mu_g$ and $\sigma^2_g$, respectively\footnote{In the case of a multiple output function $g$, i.e., $m>1$, we model each output dimension with an independent GP. We keep the notations unchanged for simplicity.}. Given a feedback control law $\Psi: \mathcal{X} \rightarrow \U$, we use $\xi_{x_0, \Psi}: \N \rightarrow \mathcal{X}$ to denote the closed-loop trajectory of~\eqref{eq:dyn} that starts from the state $x_0 \in \mathcal{X}$ and evolves under the control law $\Psi$. 

In this paper, our primary focus is on controlling the nonlinear system~\eqref{eq:dyn} with a state-feedback neural network controller $\NN: \mathcal{X} \rightarrow \U$. A $F$-layer Rectified Linear Unit (ReLU) NN is specified by composing $F$ layer functions (or just layers). A layer $l$ with $\mathfrak{i}_l$ inputs and $\mathfrak{o}_l$ outputs is specified by a weight matrix $W^{(l)} \in \R^{\mathfrak{o}_l \times \mathfrak{i}_l}$ and a bias vector $b^{(l)} \in \R^{\mathfrak{o}_l}$ as follows:
\begin{equation}
    L_{\theta^{(l)}}: z \mapsto \max\{ W^{(l)} z + b^{(l)}, 0 \}, 
\end{equation}
where the $\max$ function is taken element-wise, and $\theta^{(l)} \triangleq (W^{(l)}, b^{(l)})$ for brevity. Thus, a $F$-layer ReLU NN is specified by $F$ layer functions $\{L_{\theta^{(l)}} : l = 1, \dots, F\}$ whose input and output dimensions are composable: that is they satisfy $\mathfrak{i}_{l} = \mathfrak{o}_{l-1}: l = 2, \dots, F$. Specifically:
\begin{equation}
	\NN_\theta(x) = (L_{\theta^{(F)}} \circ L_{\theta^{(F-1)}} \circ \dots \circ L_{\theta^{(1)}})(x),
\end{equation}
where we index a ReLU NN function by a list of matrices $\theta \triangleq (\theta^{(1)}, \dots, \theta^{(F)})$. Also, it is common to allow the final layer function to omit the $\max$ function altogether, and we will be explicit about this when it is the case.

\subsection{Task and Specification}
We use $\W = \{\mathcal{X}_\goal, \mathcal{O}_1, \ldots, \mathcal{O}_o\}$ to denote a task where $\mathcal{X}_\goal \subset \mathcal{X}$  is the goal  that the system would like to reach and $\{\mathcal{O}_1, \ldots, \mathcal{O}_o\}$ with $\mathcal{O}_i \subset \mathcal{X}$ is the set of obstacles that the system would like to avoid. More formally, given a task $\W$, a safety specification $\phi_{\text{safety}}$ requires avoiding all the obstacles and a liveness specification $\phi_{\text{liveness}}$ requires reaching the goal in a bounded time horizon $H$. We use $\xi_{x_0, \Psi} \models \phi_{\text{safety}}$ and $\xi_{x_0, \Psi} \models \phi_{\text{liveness}}$ to denote a trajectory $\xi_{x_0, \Psi}$ satisfies the safety and liveness specifications, respectively, i.e.,
{\small
\begin{align*}    
    &\xi_{x_0, \Psi} \models \phi_{\text{safety}} \Longleftrightarrow \forall k \in \mathbb{N},\; \forall i \in \{\mathcal{O}_1, \ldots, \mathcal{O}_o\},\; \xi_{x_0, \Psi}(k) \not\in \mathcal{O}_i,  \\
    &\xi_{x_0, \Psi} \models \phi_{\text{liveness}} \Longleftrightarrow \exists k \in \{1,\ldots H\},\; \xi_{x_0, \Psi}(k) \in \mathcal{X}_\goal.
\end{align*}}%
Given a set of initial states $\mathcal{X}_\text{init}$, a control law $\Psi: \mathcal{X} \rightarrow \U$ satisfies a specification $\phi$ (denoted by $\Psi, \mathcal{X}_\text{init} \models \phi$) if all trajectories starting from the set $\mathcal{X}_\text{init}$ satisfy the specification, i.e., $\xi_{x, \Psi} \models \phi$, $\forall x \in \mathcal{X}_\text{init}$. Since the specifications and the satisfying set of initial states depend on the task, we explicitly add $\W$ as a superscript whenever need emphasize the dependency, such as $\phi_\text{safety}^\W$, $\phi_\text{liveness}^\W$, and $\mathcal{X}_\text{init}^\W$.

While conventional reinforcement learning focuses on training a neural network that works for one specific task, meta-RL focuses, instead, on training controllers that can work for a multitude of tasks. To formally capture this requirement, we use $\mathfrak{W}_\mathcal{X}$ to denote the set of all the tasks (corresponding to configurations of the goal and obstacles) with the goals and the obstacles be defined over the state space $\mathcal{X}$.
Though an arbitrary task such as the case of the goal is enclosed by obstacles may not be interesting, we use the set $\mathfrak{W}_\mathcal{X}$ in the statement of our problem for simplicity.

\subsection{Main Problem} 
We consider the problem of designing provably correct NN controllers for unseen tasks. Specifically, the task $\W \in \mathfrak{W}_\mathcal{X}$ is unknown during the training of the NN controller. The task $\W$ will be known only at runtime.
Therefore, our objective is to train a set (or a collection) of different ReLU NNs along with a selection algorithm that can select the correct NNs once the task $\W$ becomes available at runtime. Before presenting the problem under consideration, we introduce the following notion of NN composition.
\begin{definition}
Given a set of Neural Networks $\mathfrak{NN} = \{\NN_1, \NN_2, \ldots, \NN_m\}$ along with an activation map $\Gamma:\mathcal{X} \to \{1,\ldots,m\}$, the composed neural network $\NN_{[\mathfrak{NN},\Gamma]}$ is defined as:
$$ \NN_{[\mathfrak{NN},\Gamma]}(x) = \NN_{\Gamma(x)}(x) $$
\end{definition}
In other words, the activation map $\Gamma$ selects the index of the NN that need to be activated at a particular state $x \in \mathcal{X}$. Now, we can define the problem of interest as follows.

\begin{problem}
    \label{prob:main}
    Given the nonlinear dynamical system~\eqref{eq:dyn}. Design a NN controller $\NN$ consists of two parts: a set of ReLU NNs $\mathfrak{NN} = \{\NN_1, \NN_2, \ldots, \NN_m\}$ and a selection algorithm $\texttt{SEL}$, such that for any task $\W \in \mathfrak{W}_\mathcal{X}$, the selection algorithm $\texttt{SEL}(\W, \mathfrak{NN})$ returns a set of initial states $\mathcal{X}_\text{init}^\W \subseteq \mathcal{X}$ and an activation map $\Gamma^\W$ satisfying:
    $$ \NN_{[\mathfrak{NN},\Gamma^\W]}, \mathcal{X}_\text{init}^\W \models \phi_\text{safety}^\W \land \phi_\text{liveness}^\W.$$
\end{problem}

Indeed, it is desirable that the algorithm $\texttt{SEL}$ computes the \emph{largest} possible $\mathcal{X} _\text{init}^\W$ for the task $\W$. While computing the largest possible set can be computationally demanding, our algorithm will instead focus on finding an $\epsilon$ sub-optimal $\mathcal{X} _\text{init}^\W$. For space considerations, the quantification of the sub optimality in the computations of $\mathcal{X} _\text{init}^\W$ is omitted.

\section{FRAMEWORK}
\subsection{Overview}
Before describing our approach to solve Problems~\ref{prob:main}, we start by recalling that every $\R^n \rightarrow \R^m$ ReLU NN represents a Continuous Piece-Wise Affine (CPWA) function~\cite{pascanu2013number}. We use $\Psi_\text{CPWA}: \mathcal{X} \rightarrow \R^m$ to denote a CPWA function of the form:
\begin{equation}
    \label{eq:cpwa}
    \Psi_\text{CPWA}(x) = K_i x + b_i\quad \text{if}\ x \in \mathcal{R}_i,\ i =1, \ldots, L,
\end{equation}
where the polytopic sets $\{\mathcal{R}_1, \ldots, \mathcal{R}_L\}$ is a partition of the set $\mathcal{X}$. We call each polytopic set $\mathcal{R}_i \subset \mathcal{X}$ a linear region, and use $\mathbb{L}_{\Psi_\text{CPWA}} = \{\mathcal{R}_1, \ldots, \mathcal{R}_L\}$ to denote the set of linear regions associated with $\Psi_\text{CPWA}$. In this paper, we confine our attention to CPWA controllers (and hence neural network controllers) that are selected from a bounded polytopic set {$\mathcal{P}^{K} \times \mathcal{P}^{b} \subset \mathbb{R}^{m\times n} \times \mathbb{R}^m$}, i.e., we assume that $K_i \in \mathcal{P}^{K}$ and $b_i \in \mathcal{P}^{b}$.

To fulfill a set of \emph{infinitely} many tasks $\mathfrak{W}_\mathcal{X}$ using a \emph{finite} set of ReLU NNs $\mathfrak{NN}$, our approach is to restrict each NN in the set $\mathfrak{NN}$ to some local behavior, yet the set $\mathfrak{NN}$ captures all possible behavior of the system. We use the mathematical model of the physical system~\eqref{eq:dyn} to guide training of the NNs, as well as selecting NNs from the set $\mathfrak{NN}$ at runtime.

During training, without knowing the tasks, we train a set of ReLU NNs $\mathfrak{NN}$ using the following two steps:
\begin{itemize}
    \item Capture the closed-loop behavior of the system under \emph{all} CPWA controllers using a finite-state Markov decision process (MDP). To define the action space of this MDP, we partition the space of all CPWA controllers into a finite number of partitions. Each partition corresponds to a family of CPWA controllers. Hence, each transition in the MDP is labeled by a symbol that corresponds to a particular family of CPWA functions. The transition probabilities can then be computed using the knowledge of the model~\eqref{eq:dyn} and the Gaussian Process $\mathcal{G}\mathcal{P}(\mu_g, \sigma^2_g)$.
    We refer to this finite-state MDP as the abstract model of the system.
    
    \item Train one NN corresponds to each transition in the MDP. We refer to each of these NNs as a local NN. Let $\mathfrak{NN}$ be the set of all such local NNs. The training enforces each local NN to represent a CPWA function that belongs to the family of CPWA controllers associated with this transition. This is achieved by using the NN weight projection operator introduced in~\cite{sun2021safeRL}. Using these local NNs, we can construct the set of NN controllers $\mathfrak{NN}$.
\end{itemize}
Details of constructing the abstract model and training the local NN controllers in $\mathfrak{NN}$ are given in Section~\ref{sec:train_all_nns}.

At runtime, given an arbitrary task $\W \in \mathfrak{W}_\mathcal{X}$, the algorithm $\texttt{SEL}(\W, \mathfrak{NN})$ selects NNs from the set $\mathfrak{NN}$ to satisfy $\phi_\text{safety}^\W \land \phi_\text{liveness}^\W$:
\begin{itemize}
    \item To satisfy the safety specification $\phi_\text{safety}^\W$, the algorithm $\texttt{SEL}$ identifies a subset of safe CPWA controllers at each abstract state in the MDP. The selected NNs from the set $\mathfrak{NN}$ must correspond to one of those CPWA families that are marked as safe.
    \item For the liveness specification $\phi_\text{liveness}^\W$, the algorithm $\texttt{SEL}$ first searches for the optimal policy of the MDP using dynamic programming (DP), where the allowed transitions in the MDP are limited to those have been identified to be safe. Based on the optimal policy of the MDP, it decides which local NN in the set $\mathfrak{NN}$ should be used at each state.  
\end{itemize}

We highlight that the proposed framework above \emph{always} guarantees that the resulting NN controller satisfies the safety specification $\phi_\text{safety}^\W$ for any task $\W \in \mathfrak{W}_\mathcal{X}$, regardless the accuracy of the learned model-error using GP regression. For the liveness specification $\phi_\text{liveness}^\W$, due to the learned model-error is probabilistic, we relax Problem~\ref{prob:main} to maximize the probability of satisfying the liveness specification $\phi_\text{liveness}^\W$. We also provide a quantified bound on the probability for the NN controller to satisfy $\phi_\text{liveness}^\W$. 

\begin{figure}[h]  
  \center
  \resizebox{.4\textwidth}{!}{
    \subfloat[]{\includegraphics[width=0.17\linewidth]{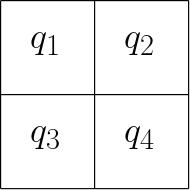}}
    \hspace{2mm}
    \subfloat[]{\includegraphics[width=0.2\linewidth]{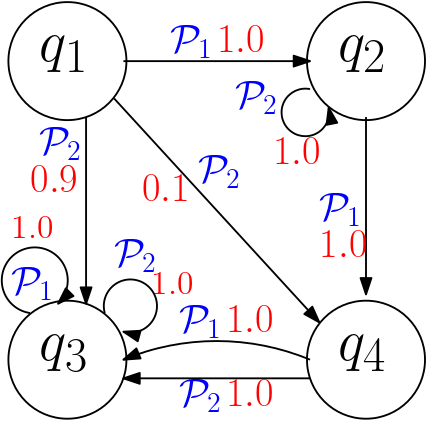}}
  }
  \caption{(a) Partition the state space into four abstract states; (b) MDP labeled with transition probabilities (red).} \vspace{-5mm}
  \label{fig:frame} 
\end{figure}

Figure~\ref{fig:frame} conceptualizes our framework. In Figure~\ref{fig:frame} (a), we partition the state space $\mathcal{X}$ into a set of abstract states $\X = \{q_1, q_2, q_3, q_4\}$ and the controller space $\p^K \times \p^b$ into a set of controller partitions $\mathbb{P} = \{\p_1, \p_2\}$. Figure~\ref{fig:frame} (b) shows the resulting MDP, with transition probabilities labeled by the side of the transitions. Then, the set $\mathfrak{NN}$ contains 9 local NNs corresponding to the 9 transitions in the MDP. 

Consider two different tasks given at runtime. Task $\W_1$ specifies that the goal $\mathcal{X}_\text{goal}$ is represented by the abstract state $q_4$ and the only obstacle is $q_2$. At state $q_1$, our selection algorithm decides to use the local network $\text{NN}_{(q_1, \p_2, q_4)}$, which corresponds to the transition from state $q_1$ to $q_4$ under partition $\p_2$. In task $\W_2$, state $q_4$ is still the goal, but there is no obstacle. For this task, our selection algorithm decides to use $\text{NN}_{(q_1, \p_1, q_2)}$ at state $q_1$ and use $\text{NN}_{(q_2, \p_1, q_4)}$ at state $q_2$. Notice that with this choice the probability of reaching the goal is $1$, which is higher than the probability $0.1$ by using $\text{NN}_{(q_1, \p_2, q_4)}$ at state $q_1$. 

In the above procedure, the set $\mathfrak{NN}$ may contain a large number of local NNs---one for every possible transition in the MDP---and need extensive training effort. To accelerate the training process, in Section~\ref{sec:transfer}, we employ ideas from transfer learning to enable the use of partially complete $\mathfrak{NN}$ to rapidly train new NN controllers, at runtime, while satisfying the same guarantees of having a complete $\mathfrak{NN}$. 

\section{PROVABLY-CORRECT TRAINING OF THE SET OF NEURAL NETWORKS $\mathfrak{NN}$}
\label{sec:train_all_nns}
\subsection{Abstract Model}
\label{subsec:abstraction}
In this section, we extend the abstract model proposed in~\cite{sun2021safeRL} by taking into account the unknown model-error $g$. Unlike the results reported in~\cite{sun2021safeRL} where the system was assume to be error-free and deterministic (and hence can be abstracted by a finite-state machine), in this paper, the dynamical model~\eqref{eq:dyn} is stochastic due to the use of GP regression to capture the error in the model. This necessitates the use of finite-state MDP to abstract the dynamics in~\eqref{eq:dyn}.


\noindent\textbf{State and Controller Space Partitioning:}
We partition the state space $\mathcal{X} \subset \R^n$ into a set of abstract states, denoted by $\X = \{q_1, \ldots, q_N\}$. Each $q_i \in \X$ is an infinity-norm ball in $\R^n$ centered around some state $x_i \in \mathcal{X}$.
The partitioning satisfies $\mathcal{X} = \bigcup_{q \in \X} q$, and $\Int(q_i) \cap \Int(q_j) = \emptyset$ if $i \neq j$. With an abuse of notation, $q$ denotes both an abstract state, i.e., $q \in \X$, and a subset of states, i.e., $q \subset \mathcal{X}$. Since we construct the abstract model before knowing the tasks, the state space $\mathcal{X}$ does not contain any obstacle or goal.

Similarly, we partition the controller space into polytopic subsets. For simplicity of notation, we define the set of parameters $\p^{K \times b} \subset \R^{m \times (n+1)}$ be a polytope that combines $\p^K \subset \R^{m \times n}$ and $\p^b \subset \R^m$. With some abuse of notation, we use $K_i(x)$ with a single parameter $K_i \in \p^{K \times b}$ to denote $K'_i x_i + b'_i$ with the pair $(K'_i,b'_i) = K_i$. The controller space  $\p^{K \times b}$ is discretized into a collection of polytopic subsets in $\R^{m \times (n+1)}$, denoted by $\mathbb{P} = \{\p_1, \ldots, \p_M\}$. Each $\p_i$ is an infinity-norm ball centered around some $K_i$ 
such that $\p^{K \times b} = \bigcup_{\p \in \mathbb{P}} \p$, and $\Int(\p_i) \cap \Int(\p_j) = \emptyset$ if $i \neq j$. We call each of the subsets $\p_i \in \mathbb{P}$ a controller partition. Each controller partition $\p \in \mathbb{P}$ represents a subset of CPWA functions, by restricting parameters $K_i$ in a CPWA function to take values from $\p$. 


\noindent\textbf{MDP Transitions:}
Next, we compute the set of all allowable transitions in the MDP. To that end, we define the posterior of an abstract state $q$ under a controller partition $\p$ be the set of states that can be reached in one step from states $x \in q$ by using affine state feedback controllers with parameters $K \in \p$ under the dynamical model~\eqref{eq:dyn} as follows:
%
\begin{equation}
    \label{eq:post}
    \Post(q, \p) \triangleq \{h(x, K(x)) \in \R^n\; |\; x \in q, K \in \p\} \oplus \mathcal{D},
\end{equation}
where $\mathcal{D} \subset \R^n$ is defined in Section~\ref{subsec:model} as the bound of the model-error $g$.
Indeed, computing the exact posterior for a nonlinear system is computationally expensive, and hence we rely on over-approximation $\widehat{\Post}(q,\p)$ instead.
%
Furthermore, let $\Next(q, \p)$ be the set of abstract states that have overlap with $\Post(q, \p)$.
\begin{equation}
    \label{eq:next}
    \Next(q, \p) \triangleq \{q^\prime \in \X \; | \; q^\prime \cap \widehat{\Post}(q, \p) \neq \emptyset\}.
\end{equation}
The transitions in the MDP can now be constructed using the information in $\Next(q, \p)$. That is, a transition from state $q$ to state $q'$ with label $\p$ is allowed in the MDP if and only if $q' \in \Next(q, \p)$.

\noindent\textbf{Transition Probability:}
The final step is to compute the transition probabilities associated with each of the transitions constructed in the previous step. We define transition probabilities based on representative points in abstract states and controller partitions. Specifically, we choose the representative points to be the centers (recall that both $q$ and $\p$ are infinity-norm balls and hence their centers are well defined). Let $\ct_\X: \X \rightarrow \mathcal{X}$ map an abstract state $q \in \X$ to its center 
and $\ct_\mathbb{P}: \mathbb{P} \rightarrow \p^{K \times b}$ map a controller partition $\p \in \mathbb{P}$ to the matrix $\ct_\mathbb{P}(\p) \in \p^{K \times b}$, which is the center of $\p$. Furthermore, we use $\abs_\X: \mathcal{X} \rightarrow \X$ to denote the map from a state $x \in \mathcal{X}$ to the abstract state that contains $x$, i.e., $x \in \abs_\X(x) \in \X$, and similarly, the map $\abs_\mathbb{P}: \p^{K \times b} \rightarrow \mathbb{P}$ satisfies $K \in \abs_\mathbb{P}(K) \in \mathbb{P}$ for any $K \in \p^{K \times b}$. 

Given the dynamical system~\eqref{eq:dyn} with the model-error $g$ learned by a GP regression model $\mathcal{G}\mathcal{P}(\mu_g, \sigma^2_g)$, let $t: \mathcal{X} \times \mathcal{X} \times \U \rightarrow [0, 1]$ be the corresponding conditional stochastic kernel. Specifically, given the current state $x \in \mathcal{X}$ and input $u \in \mathcal{U}$, the distribution $t(\cdot|x,u)$ is given by the Gaussian distribution $\mathcal{N}(f(x,u)+\mu_g(x,u), \sigma^2_g(x,u))$. For any set $\mathcal{A} \in \mathcal{B}(\mathcal{X})$ and any $k \in \N$, the probability of reaching the set $\mathcal{A}$ in one step from state $x^{(k)}$ with input $u^{(k)}$ is given by:
\begin{equation}
    \label{eq:integrate_A}
    \pr(x^{(k+1)} \in \mathcal{A} | x^{(k)}, u^{(k)}) = \int_\mathcal{A} T(dx^{(k+1)} | x^{(k)}, u^{(k)})
\end{equation}
where we use the notation $T(dx^\prime|x,u) \triangleq t(x^\prime|x,u) \mu(dx^\prime)$. This integral can be easily computed since $t(\cdot | x, u)$ is a Gaussian distribution\footnote{In the case of a multiple output function $g$, i.e., $m>1$, each dimension can be integrated independently.}.

With above notations, we define our abstract model as follows:
\begin{definition}
    \label{def:finite_mdp}
    The abstract model of~\eqref{eq:dyn} is a finite MDP $\hat{\Sigma}$ defined as a tuple $\hat{\Sigma} = (\X, \mathbb{P}, \hat{T})$, where:
    \begin{itemize}
        \item The state space is the set of abstract states $\X$;
        \item The set of controls at each state is given by the set of controller partitions $\mathbb{P}$;
        \item The transition probability from state $q \in \X$ to $q^\prime \in \X$ with label $\p \in \mathbb{P}$ is given by:
        \begin{align}
            \hat{T}(q^\prime|q,\p)=
            \begin{cases}
                \int_{q^\prime} t(dx^\prime|z, \kappa(z))\ &\text{if } q^\prime \in \Next(q, \p) \\
                0\ &\text{else} \notag
            \end{cases}
        \end{align}
        where $z=\ct_\X(q)$, $\kappa=\ct_\mathbb{P}(\p)$.
    \end{itemize}
\end{definition}

\subsection{Train Local NNs with Weight Projection}
Once the abstract model is computed, the next step is to train the set of local neural networks $\mathfrak{NN}$ without the knowledge of the tasks. In order to capture the closed-loop behavior of the system under \emph{all} possible CPWA controllers, we train one local NN corresponding to each transition (with non-zero transition probability) in the MDP $\hat{\Sigma}$. 
Algorithm~\ref{alg:offline-train} outlines training of all the local NNs. 
We use $\NN_{(q, \p, q^\prime)}$ to denote the local NN corresponding to the transition in the MDP $\hat{\Sigma}$ from abstract state $q \in \X$ to $q^\prime \in \X$ under controller partition $\p \in \mathbb{P}$. 

We train each local network $\NN_{(q, \p, q^\prime)}$ using Proximal Policy Optimization (PPO)~\cite{ppo} (line~\ref{line:ppo} in Algorithm~\ref{alg:offline-train}). 
While choosing the reward function in reinforcement learning is often challenging, 
our algorithm enjoys a straightforward yet efficient formulation of reward functions. To be specific, for a local network $\NN_{(q, \p, q^\prime)}$, let $\kappa = \ct_\mathbb{P}(\p)$ and $w_1, w_2 \in \R$ be pre-specified weights, our reward function encourages moving towards the state $q^\prime$ with controllers chosen from the partition $\p$: 
\begin{align*}
    &r(x, u) = \\
    &\begin{cases}
        -w_2 \norm{u - \kappa(x)}, \qquad \qquad \qquad \text{if } h(x, u) + \mu_g(x, u) \in q^\prime \\~\\
        -w_1 \norm{h(x, u) + \mu_g(x, u) - \ct_\X(q^\prime)} - w_2 \norm{u - \kappa(x)} \\ 
        \qquad \qquad \qquad \qquad \qquad \qquad  \quad \text{otherwise} \notag
    \end{cases}
\end{align*}
where $\mu_g$ is the posterior mean function from the GP regression.
With this dynamical model, PPO can efficiently explore the workspace without running the real agent. 

The training of local networks $\NN_{(q, \p, q^\prime)}$ is followed by applying a NN weight projection operator \texttt{Project} introduced in~\cite{sun2021safeRL}. 
Given a neural network $\NN$ and a controller partition $\p$, this projection operator ensures that: 
$$\texttt{Project}(\NN, \p) \in \p. $$ 
In other words, this projection operator forces that $\NN$ can only give rise to one of the CPWA functions that belong to the controller partition $\p$.
We refer readers to~\cite{sun2021safeRL} for more details on the NN weight projection. Algorithm~\ref{alg:offline-train} summarizes the discussion in this subsection.

\begin{algorithm}
    \caption{\textsc{TRAIN-LOCAL-NNs} ($\hat{\Sigma}$)}
    \label{alg:offline-train}
    {\small
    \begin{algorithmic}[1]
        \STATE $\mathfrak{NN} = \{\}$
        \FOR{$q \in \X$}
            \FOR{$\p \in \mathbb{P}$}
                \FOR{$q^\prime \in \Next(q, \p)$} \label{line:next}
                    \STATE $\NN_{(q, \p, q^\prime)} = \texttt{PPO}(q, \p, q^\prime,h, \mu_g)$ \label{line:ppo}
                    \STATE $\NN_{(q, \p, q^\prime)} = \texttt{Project}(\NN_{(q, \p, q^\prime)}, \p)$ \label{line:project}
                    \STATE $\mathfrak{NN} = \mathfrak{NN} \; \cup \; \{\NN_{(q, \p, q^\prime)} \}$
                \ENDFOR
            \ENDFOR
        \ENDFOR
        \STATE \textbf{Return} $\mathfrak{NN}$
    \end{algorithmic}  
    }
\end{algorithm}

\section{THE SELECTION ALGORITHM $\texttt{SEL}(\W,\mathfrak{NN})$}
\label{sec:algorithm}
In this section, we present our selection algorithm $\texttt{SEL}(\W,\mathfrak{NN})$ which is used at runtime when an arbitrary task $\W \in \mathfrak{W}_\mathcal{X}$ is given. The $\texttt{SEL}(\W,\mathfrak{NN})$ algorithm assigns one local NN in the set $\mathfrak{NN}$ to each abstract state in order to satisfy the safety and liveness specification $\phi_\text{safety} \wedge \phi_\text{liveness}$. Our approach is to first exclude all transitions in the MDP that can lead to violation of $\phi_\text{safety}$, followed by selecting the optimal solution from the remaining transitions in the MDP. More details are given below.


\subsection{Exclude Unsafe Transitions using Backtracking}
Given a task $\W \in \mathfrak{W}_\mathcal{X}$ that specifies a set of obstacles $\{\mathcal{O}_1, \ldots, \mathcal{O}_o\}$ and a goal $\mathcal{X}_\goal$, we use $\X_\text{obst}$ to denote the subset of abstract states that intersect the obstacles, i.e., $\X_\text{obst} = \{q \in \X | \exists i \in \{1, \ldots, o\}, q \bigcap \mathcal{O}_i \neq \emptyset\}$, and use $\X_\text{goal}$ to denote the subset of abstract states contained in the goal, i.e., $\X_\text{goal} = \{q \in \X | q \subseteq \mathcal{X}_\text{goal}\}$.

Algorithm~\ref{alg:backtrack_safety} computes the set of safe states and safe controller partitions using an iterative backward procedure introduced in~\cite{sun2021safeRL}. With the set of unsafe states initialized to be the obstacles (line~\ref{line:initialize_unsafe} in Algorithm~\ref{alg:backtrack_safety}), the algorithm backtracks unsafe states until a fixed point is reached, i.e., it can not find new unsafe states (line~\ref{line:iter_begin}-\ref{line:iter_end} in Algorithm~\ref{alg:backtrack_safety}). The set of safe initial states $\mathcal{X}^\W_\text{init}$ is the union of all the abstract states that are identified to be safe (line~\ref{line:safe_init} in Algorithm~\ref{alg:backtrack_safety}). Furthermore, it computes the function $P^\W_\safe: \X^\W_\text{safe} \rightarrow 2^\mathbb{P}$, which assigns a set of safe controller partitions $P^\W_\safe(q) \subseteq \mathbb{P}$ at each abstract state $q \in \X^\W_\text{safe}$. Again, we use the superscript $\W$ to emphasize the dependency of $\mathcal{X}^\W_\text{init}$, $\X^\W_\safe$ and $P_\safe^\W$ on the task $\W$.

\begin{algorithm}
    \caption{\textsc{BACKTRACK-SAFETY} ($\hat{\Sigma}$, $\W$)}
    \label{alg:backtrack_safety}
    {\small
    \begin{algorithmic}[1]
        \STATE $\X_\text{unsafe}^0 = \emptyset$, $\X_\text{unsafe}^1 = \X_\text{obst}$, $k=1$ \label{line:initialize_unsafe}
        \WHILE{$\X_\text{unsafe}^k \neq \X_\text{unsafe}^{k-1}$} \label{line:iter_begin}
            \STATE $\X_\text{unsafe}^{k+1} =\hspace{-1mm} \{q \in \X | \forall \p \in \mathbb{P}: \Next(q, \p) \cap \X_\text{unsafe}^k \neq \emptyset\} \cup \X_\text{unsafe}^k$ 
            \STATE $k = k+1$ \label{line:iter_end}
        \ENDWHILE
        \STATE $\X_\text{safe}^\W = \X^\W \setminus \X_\text{unsafe}^k$
        \STATE $\mathcal{X}_\text{init}^\W = \bigcup_{q \in \X^\W_\safe} q$ \label{line:safe_init}
        \FOR{$q \in \X_\text{safe}^\W$}
            \STATE $P_{\text{safe}}^\W(q) = \{\p \in \mathbb{P}\; |\; \Next(q, \p) \cap \X_\text{unsafe}^k = \emptyset\}$
        \ENDFOR
        \STATE \textbf{Return} $\mathcal{X}^\W_\text{init}$, $\X_\text{safe}^\W$, $\{P_{\text{safe}}^\W(q)\}_{q \in \X^\W_\text{safe}}$
    \end{algorithmic}  
    }
\end{algorithm}

\subsection{Assign Controller Partition by Solving MDP}
Once the set of safe controller partitions $P^\W_\safe(q)$ is computed, the next step is to assign one controller partition in $P^\W_\safe(q)$ to each abstract state $q$. In particular, we consider the problem of solving the optimal policy for the MDP $\hat{\Sigma}$ with states and controls limited to the set of safe abstract states $\X^\W_\safe$ and the set of safe controller partitions $P^\W_\safe(q)$ at $q \in \X^\W_\safe$, respectively. Since we are interested in maximizing the probability of satisfying the liveness specification $\phi_\text{liveness}$, let the optimal value function $\hat{V}^*_k: \X^\W_\safe \rightarrow [0, 1]$ map an abstract state $q \in \X^\W_\safe$ to the maximum probability of reaching the goal in $H-k$ steps from $q$. Using this notation, $\hat{V}^*_0(q)$ is then the maximum probability of satisfying the liveness specification $\phi_\text{liveness}$. 
The optimal value functions can be solved by the following Dynamic Programming (DP) recursion~\cite{abate2013hscc}:
\begin{align}
    \hat{V}_k(q, \p) &=\!\!\mathbf{1}_{\X_\text{goal}}(q) \!+\! \mathbf{1}_{\X_\safe \setminus \X_\text{goal}}(q)\!\! \sum_{q^\prime \in \X^\W_\safe} \!\! \hat{V}_{k+1}^* (q^\prime) \hat{T}(q^\prime | q, \p) \label{eq:abst_Q} \\
    \hat{V}_k^*(q) &= \max_{\p \in P^\W_\safe(q)} \hat{V}_k(q, \p) \label{eq:abst_V}
\end{align} 
with the initial condition $\hat{V}_H^* =  \mathbf{1}_{\X_\text{goal}}$, where $k = H, \ldots, 0$.

Algorithm~\ref{alg:dp_liveness} solves the optimal policy for the MDP $\hat{\Sigma}$ using the Dynamic Programming (DP) recursion~\eqref{eq:abst_Q}-\eqref{eq:abst_V}. At time step $k$, the optimal controller partition $\p^*$ at state $q$ is given by the maximizer of $\hat{V}_k(q, \p)$ (line~\ref{line:p_star} in Algorithm~\ref{alg:dp_liveness}). The last step is to assign a corresponding neural network to be used at all the states $x \in q$ for each $q \in \X_\safe$. To that end, the activation map $\Gamma_{k,\text{abs}}^{\W}$ assigns the neural network indexed by $(q, \p^*, q^{\prime*})$ to the abstract state $q$, where $q^{\prime*}$ maximizes the transition probability $\hat{T}(q^\prime|q, \p^*)$ (line~\ref{line:q_prime_star}-\ref{line:assign} in Algorithm~\ref{alg:dp_liveness}). While the activation map $\Gamma_{k,\text{abs}}^{\W}$ assigns a neural network index to the abstract state $q$, we can directly get the activation map to the actual state $x \in \mathcal{X}$ as:
$$ \Gamma_k^\W(x) = \Gamma_{k,\text{abs}}^{\W}(\abs_\X(x)).$$
In other words, given the state of the system $x$, we first compute the corresponding abstract state $\abs_\X(x)$, and use the corresponding neural network assigned to this abstract state to control the system. Note that, unlike the definition of the activation map $\Gamma^\W$ in Problem~\ref{prob:main}, the activation map obtained here is time-varying as captured by the subscript $k, k = 0, \ldots, H$. This reflects the nature of the optimal solution computed by the DP regression~\eqref{eq:abst_Q}-\eqref{eq:abst_V}.

\begin{algorithm}
    \caption{\textsc{DP-LIVENESS} ($\hat{\Sigma}$, $\X_\safe^\W$, $\{P_{\text{safe}}^\W(q)\}_{q \in \X^\W_\text{safe}})$}
    \label{alg:dp_liveness}
    {\small
    \begin{algorithmic}[1]
        \FOR{$q \in \X^\W_\safe$}
            \STATE $\hat{V}_H^*(q) =  \mathbf{1}_{\X_\text{goal}} (q)$
        \ENDFOR
        \STATE $k = H-1$
        \WHILE{$k \geq 0$}
            \FOR{$q \in \X^\W_\safe$}
                \STATE $\hat{V}_k(q, \p) = \mathbf{1}_{\X_\text{goal}}(q) \; + $ \\
                \hspace{14mm}$\mathbf{1}_{\X_\safe \setminus \X_\text{goal}}(q) \sum_{q^\prime \in \X^\W_\safe} \hat{V}_{k+1}^* (q^\prime) \hat{T}(q^\prime | q, \p)$
                \STATE $\hat{V}_k^*(q)  = \underset{\p \in P^\W_\safe(q)}{\max} \hat{V}_k(q, \p)$
                \STATE $\p^* = \underset{\p \in P^\W_\safe(q)}{\text{argmax}} \;\hat{V}_k(q, \p)$ \label{line:p_star}
                \STATE $q^{\prime*} = \underset{q^\prime \in \X^\W_\safe}{\text{argmax}} \;\; \hat{T}(q^\prime|q, \p^*)$ \label{line:q_prime_star}
                \STATE $\Gamma_{k,\text{abs}}^{\W}(q) = (q, \p^*, q^{\prime*})$ \label{line:assign}
                
            \ENDFOR
            \STATE $k = k-1$
        \ENDWHILE
        \STATE \textbf{Return} $\Gamma_{k,\text{abs}}^{\W}$ 
    \end{algorithmic}  
    }
\end{algorithm} 
The $\mathcal{X}^\W_\text{init}$ computed by Algorithm~\ref{alg:backtrack_safety} along with the selection map $\Gamma_{k,\text{abs}}^{\W}$ returned by Algorithm~\ref{alg:dp_liveness} constitutes the $\texttt{SEL}(\W, \mathfrak{NN})$ algorithm.

\section{THEORETICAL GUARANTEES}
\label{sec:guarantee}
In this section, we study the theoretical guarantees of the proposed solution. We analyze the guarantees of satisfying $\phi_{\text{safety}}$ and $\phi_\text{liveness}$ separately. 

\subsection{Safety Guarantee}
The following theorem summarizes the safety guarantees for our solution.
\begin{theorem}
    \label{thm:safety}
    Consider the dynamical model~\eqref{eq:dyn}. Let the NN controller $\NN$ consists of two parts: the set of local neural networks $\mathfrak{NN}$ trained by Algorithm~\ref{alg:offline-train} and the selection algorithm $\texttt{SEL}$ defined by Algorithm~\ref{alg:backtrack_safety} and Algorithm~\ref{alg:dp_liveness}. 
    For any task $\W \in \mathfrak{W}_\mathcal{X}$, consider  the set of initial conditions $\mathcal{X}^\W_\text{init}$ and the activation map $\Gamma_k^\W$ computed by $\texttt{SEL}(\W,\mathfrak{NN})$, the following holds:
    $\NN_{[\mathfrak{NN}, \Gamma_k^\W]}, \mathcal{X}^\W_\text{init} \models \phi^\W_\text{safe}$.  
\end{theorem}
The proof of Theorem~\ref{thm:safety} follows the same argument of the error-free case presented in~\cite{sun2021safeRL} and hence is omitted for brevity. 
In particular, Theorem 4.2 in~\cite{sun2021safeRL} shows that at safe abstract states $q \in \X_\safe$, any feedback CPWA controller with $K_i$ chosen from $\p \in P_\safe(q)$ is guaranteed to be safe. Furthermore, Theorem 4.4 in~\cite{sun2021safeRL} shows that the NN weight projection operator $\texttt{Project}$ ensures that the local NNs at $q \in \X_\safe$ only give rise to the feedback CPWA controllers with $K_i \in \p$ for some $\p \in P_\safe(q)$.

To take into account the model-error $g$, the posterior in~\eqref{eq:post} is inflated with the error bound $\mathcal{D}$. Hence,
Algorithm~\ref{alg:backtrack_safety} provides the same safety guarantee, regardless the accuracy of the learned model-error by GP regression. With the NN weight projection in the training of local NNs (line~\ref{line:project} in Algorithm~\ref{alg:offline-train}), the resulting NN controller is guaranteed to be safe for any task $\W \in \mathfrak{W}_\mathcal{X}$.

\subsection{Probabilistic Optimality Guarantee}
Due to the unknown model-error $g$, which is learned by GP regression, the liveness specification $\phi_\text{liveness}$ may not be always satisfied. However, in this subsection, we provide a bound on the probability for the trained NN controller to satisfy $\phi_\text{liveness}$. Intuitively, this bound tells how close is the NN controller to the optimal controller, which maximizes the probability of satisfying $\phi_\text{liveness}$. 

By replacing the model-error $g$ in~\eqref{eq:dyn} using the GP regression model $\mathcal{G}\mathcal{P}(\mu_g, \sigma^2_g)$, we consider the stochastic system $x^{(k+1)} = f(x^{(k)}, u^{(k)}) + \hat{g}(x^{(k)}, u^{(k)})$, where $\hat{g}(x^{(k)}, u^{(k)}) \sim \mathcal{N}(\mu_g(x^{(k)}, u^{(k)}), \sigma^2_g(x^{(k)}, u^{(k)}))$. Given an arbitrary task $\W \in \mathfrak{W}_\mathcal{X}$, we use $\Sigma$ to denote the embedded MDP corresponding to this stochastic system, with states and controls limited to the subspace that has been identified to be safe (see Algorithm~\ref{alg:backtrack_safety})\footnote{Since the task $\W \in \mathfrak{W}_\mathcal{X}$ is fixed when comparing the NN controller and the optimal controller, we drop the superscript $\W$ in this subsection.}. Specifically, we define the continuous MDP as a tuple $\Sigma = (\mathcal{X}_\safe, \{\U_\safe(x)\}_{x \in \mathcal{X}_\safe}, \U_\safe, T)$, where:
\begin{itemize}
    \item The state space is the set of safe states $\mathcal{X}_\safe = \bigcup_{q \in \X_\safe} q \subseteq \mathcal{X} \subset \R^n$;
    \item The available controls at each state $x \in \mathcal{X}_\safe$ are given by the feedback CPWA controllers with $K_i$ chosen from the safe controller partitions, i.e., $\U_\safe(x) = \bigcup_{\p \in P_\safe(\abs(x))} \{K(x)\; |\; K \in \p\}$;
    \item The set of controls is $\U_\safe = \bigcup_{x \in \mathcal{X}_\safe} \U_\safe(x)$;
    \item The conditional stochastic kernel $T$ follows the same definition in Section~\ref{subsec:abstraction}.  
\end{itemize}

We first consider the optimal controller for the system $\Sigma$ in terms of maximizing the probability of satisfying the liveness specification $\phi_\text{liveness}$. Similar to the finite-state MDP $\hat{\Sigma}$, let the optimal value function $V_k^*: \mathcal{X}_\safe \rightarrow [0,1]$ map a state $x \in \mathcal{X}_\safe$ to the maximum probability of reaching the goal in $H-k$ steps from $x$. Let $\mathcal{X}_\safe^\prime = \mathcal{X}_\safe \setminus \mathcal{X}_\text{goal}$, the optimal value functions can be solved through DP recursion~\cite{abate2013hscc}:
\begin{align}
        V_k(x, u) &= \mathbf{1}_{\mathcal{X}_\goal}(x) + \mathbf{1}_{\mathcal{X}_\safe^\prime}(x) \int_{\mathcal{X}_\safe} V_{k+1}^* (x^\prime) T(dx^\prime | x, u) \label{eq:concrete_Q} \\
        V_k^*(x) &= \sup_{u \in \U_\safe(x)} V_k(x, u) \label{eq:concrete_V}
\end{align}
with the initial condition $V_H^* =  \mathbf{1}_{\mathcal{X}_\goal}$, where $k = H, \ldots, 0$. In the following, we use the DP recursion~\eqref{eq:concrete_Q}-\eqref{eq:concrete_V} to bound the optimality of NN controllers without explicitly solving them, which is intractable due to the continuous state and input space.

The probability for the NN controller to satisfy the liveness specification $\phi_\text{liveness}$ is given by the value function $V_k^\NN: \mathcal{X}_\safe \rightarrow [0,1]$, which maps a state $x \in \mathcal{X}_\safe$ to the probability of reaching the goal in $H-k$ steps from the state $x$ under the controller $\NN$:
\begin{equation*}
    V_k^\NN (x) \triangleq \pr(\exists k^\prime \in \{k, \ldots, H\},\; \xi_{x, \NN}(k^\prime) \in \mathcal{X}_\goal).
\end{equation*}
Similarly, $V_k^\NN (x)$ can be solved through the DP recursion:
\begin{equation}
    V_k^\NN(x) \!=\! \mathbf{1}_{\mathcal{X}_\goal}(x) + \mathbf{1}_{\mathcal{X}_\safe^\prime}\!\!(x)\!\!\! \int_{\mathcal{X}_\safe} \hspace{-5mm} V_{k+1}^\NN (x^\prime) T(dx^\prime | x, \NN(x)) \label{eq:nn_V}
\end{equation}
with the initial condition $V_H^\NN = \mathbf{1}_{\mathcal{X}_\goal}$, where $k = H, \ldots, 0$. 

With the above notations, the difference between the value functions $V_k^\NN$ and $V_k^*$ measures the optimality of the NN controller $\NN$ by comparing it with the optimal controller. The following theorem provides the upper bound on this difference. When $k=0$, it upper bounds the difference between the probability of satisfying the liveness specification $\phi_\text{livenss}$ using the NN controller and the maximum probability that can be achieved. 
\begin{theorem}
    \label{thm:optimality}
    Let $V_k^\NN$ and $V_k^*$ be the functions defined above. For any $x \in \mathcal{X}_\safe$ it holds that
    \begin{equation}
        |V^\NN_k(x) - V^*_k(x)| \leq (H-k) (\Delta^\NN + \Delta^*)
    \end{equation} where
    \begin{align*}
    \Delta^\NN &= \max_{1 \leq i \leq N^\prime} (\Lambda_i \delta_q + \Gamma_i L_i \delta_q + \sqrt{m(n+1)} \mathcal{L}_\mathcal{X} \Gamma_i \delta_\p) \\
    \Delta^* &= \max_{1 \leq i \leq N^\prime} (\Lambda_i \delta_q + \Gamma_i \mathcal{L}_\p \delta_q + 2 \sqrt{m(n+1)} \mathcal{L}_\mathcal{X} \Gamma_i \delta_\p)
    \end{align*} 
    and the constants are defined as follows:
    the number of safe abstract states $N^\prime = |\X_\safe|$, grid size $\delta_q = \underset{q \in \X_\safe}{\max}\ \underset{x, x^\prime \in q}{\max} \norm{x - x^\prime}$, and $\delta_\p = \underset{\p \in \mathbb{P}}{\max}\ \underset{K, K^\prime \in \p}{\max} \norm{K - K^\prime}_{\max}$. Furthermore, $\underset{x \in \mathcal{X}_\safe}{\sup} \norm{x} \leq \mathcal{L}_\mathcal{X}$, $\underset{K \in \p^{K \times b}}{\sup} \norm{K} \leq \mathcal{L}_\mathcal{\p}$, and $L_i$ is the Lipschitz constant of an arbitrary local NN corresponding to a transition leaving $q_i \in \X_\safe$:
    $$\forall x, x^\prime \in q_i,\ \norm{\NN_{(q_i, \p, q^\prime)}(x) - \NN_{(q_i, \p, q^\prime)}(x^\prime)} \leq L_i \norm{x - x^\prime}$$
    for any $\p \in P_\safe(q_i)$ and $q^\prime \in \Next(q_i, \p)$. Finally, $\Lambda_i = \int_{\mathcal{X}_\safe} \lambda_i(y) \mu(dy)$ and $\Gamma_i = \int_{\mathcal{X}_\safe} \gamma_i(y) \mu(dy)$, where $\lambda_i(y)$ and $\gamma_i(y)$ are the Lipschitz constants of the stochastic kernel at abstract state $q_i \in \X$, i.e., $\forall y \in \mathcal{X}$:
    \begin{align*}
    &|t(y|x^\prime, u) - t(y|x, u)| \leq \lambda_i (y) \norm{x^\prime - x}, \forall x, x^\prime \in q_i,  u \in \U \\
    &|t(y|x, u^\prime) - t(y|x, u)| \leq \gamma_i (y) \norm{u^\prime - u}, \forall u, u^\prime \in \U, x \in q_i.
    \end{align*}
\end{theorem}

Notice that the upper bound in Theorem~\ref{thm:optimality} can be arbitrarily small when the grid size $\delta_\mathcal{X}$ and $\delta_\p$ approach zero. Also, since the Lipschitz constants $\Lambda_i$ and $\Gamma_i$ depend on the abstract states $q_i \in \X$, one can reduce the Lipschitz constants by tuning the grid size. Please see the appendix for the proof of Theorem~\ref{thm:optimality}.



\section{ACCELERATE BY TRANSFER LEARNING}
\label{sec:transfer}
In Algorithm~\ref{alg:offline-train}, the set $\mathfrak{NN}$ consists of the local NNs corresponding to all the transitions in the abstract model $\hat{\Sigma}$, and this may require to train a large number of local NNs. In this section, we accelerate the training process using transfer learning. Since each local NN is associated with a transition in the MDP $\hat{\Sigma}$, distance between local NNs can be easily defined based on their associated transitions. Specifically, given two transitions $(q_1, \p_1, q^\prime_1)$ and $(q, \p, q^\prime)$ in the MDP $\hat{\Sigma}$, with pre-specified weights $\alpha_1, \alpha_2, \alpha_3 \in \R$, we use the following metric to measure the distance between them:
\begin{align}
    \text{dist}((q_1, \p_1, q^\prime_1), (q, \p, q^\prime)) = \alpha_1 \norm{\ct_\X(q_1) - \ct_\X(q)}  \notag \\  
    \alpha_2 \norm{\ct_\X(q^\prime_1) - \ct_\X(q^\prime)} + \alpha_3 \norm{\ct_\mathbb{P}(\p_1) - \ct_\mathbb{P}(\p)}_{\max}. \label{eq:nn_dist}
\end{align}

Instead of training the set of all the local networks $\mathfrak{NN}$, we only train a subset of networks $\mathfrak{NN}_\text{part} \subset \mathfrak{NN}$. In particular, we assume to have access to some tasks $\W \in \mathfrak{W}_\mathcal{X}$ for training, and use the trained NN controller for unseen tasks $\W_* \in \mathfrak{W}_\mathcal{X}$ at runtime. Given a training task $\W$, after computing the set of safe abstracts states $\X^\W_\safe$ (Algorithm~\ref{alg:backtrack_safety}) and the activation map $\Gamma_k^\W$ (Algorithm~\ref{alg:dp_liveness}), Algorithm~\ref{alg:online-train} trains one local NN associated to each abstract state $q \in \X^\W_\safe$. This procedure can be called for multiple training tasks and the set of trained local NNs is given by $\mathfrak{NN}_\text{part}$. 

At runtime, for an unseen task $\W_* \in \mathfrak{W}_\mathcal{X}$, in the case of a local NN is needed but not trained, Algorithm~\ref{alg:online-run} employs transfer learning to train the local NN online. By initializing the missing NN with the closest NN based on the distance metric~\eqref{eq:nn_dist} (line~\ref{line:close_nn}-\ref{line:init} in Algorithm~\ref{alg:online-run}), the missing NN can be trained \emph{at runtime} by PPO with only a few episodes (line~\ref{line:ppo_update} in Algorithm~\ref{alg:online-run}). Thanks to the NN weight projection operator, the resulting NN controller enjoys the same safety and optimality guarantees presented in Section~\ref{sec:guarantee} (line~\ref{line:transfer_project} in Algorithm~\ref{alg:online-run}).

\begin{algorithm}
    \caption{\textsc{TRAIN-TRANSFER} ($\hat{\Sigma}$, $\X_\text{safe}^\W$, $\Gamma^\W_k$)}
    \label{alg:online-train}
    {\small
    \begin{algorithmic}[1]
        \STATE $\mathfrak{NN}_\text{part} = \{\}$
        \FOR{$q \in \X^\W_\text{safe}$}
            \STATE $(q, \p, q^\prime) = \Gamma^\W_0(q)$
            \STATE $\NN_{(q, \p, q^\prime)} = \texttt{PPO}(q, \p, q^\prime, h, \mu_g)$ 
            \STATE $\NN_{(q, \p, q^\prime)} = \texttt{Project}(\NN_{(q, \p, q^\prime)}, \p)$
            \STATE $\mathfrak{NN}_\text{part} = \mathfrak{NN}_\text{part} \bigcup \{\NN_{(q, \p, q^\prime)}\}$
        \ENDFOR
        \STATE \textbf{Return} $\mathfrak{NN}_\text{part}$
    \end{algorithmic}   
    }
\end{algorithm}

\begin{algorithm}
    \caption{\textsc{RUN-NN-CONTROLLER-TRANSFER} \\($x^{(0)}$, $\hat{\Sigma}$, $\mathfrak{NN}_\text{part}$, $\Gamma^\W_k$)}
    \label{alg:online-run}
    {\small
    \begin{algorithmic}[1]
        \STATE $k = 0$
        \WHILE{$x^{(k)} \not \in \mathcal{X}_\text{goal}$}
            \STATE $q = \abs_\X(x^{(k)})$
            \STATE $(q, \p, q^\prime) = \Gamma^{\W_*}_k(q)$ 
            \IF{$\NN_{(q, \p, q^\prime)} \not \in \mathfrak{NN}_\text{part}$}
                \STATE $(q^*, \p^*, q^{\prime*}) = \hspace{-5mm}\underset{\NN_{(q_1, \p_1, q^\prime_1)} \in \mathfrak{NN}_\text{part}}{\text{argmin}} \hspace{-5mm} \text{dist}((q_1, \p_1, q^\prime_1), (q, \p, q^\prime))$ \label{line:close_nn}
                \STATE $\NN_{(q, \p, q^\prime)} = \text{initialize}(\NN_{(q^*, \p^*, q^{\prime*})})$ \label{line:init}
                \STATE $\NN_{(q, \p, q^\prime)} = \text{PPO\_update}(q, \p, q^\prime, h, \mu_g)$ \label{line:ppo_update}
                \STATE $\NN_{(q, \p, q^\prime)} = \texttt{Project}(\NN_{(q, \p, q^\prime)}, \p)$ \label{line:transfer_project}
            \ENDIF    
            \STATE $u = \NN_{(q, \p, q^\prime)}(x^{(k)})$
            \STATE $k = k+1$
            \STATE Apply input $u$, observe the new state $x^{(k)}$
        \ENDWHILE
    \end{algorithmic}  
    }
\end{algorithm}

\section{RESULTS}  
We implemented the proposed framework and show simulation results in this section. All experiments were executed on an Intel Core i9 2.4-GHz processor with 32 GB of memory. 

Consider a wheeled robot with the state vector $x = [\zeta_x, \zeta_y, \theta]^\top \in \mathcal{X} \subset \R^3$, where $\zeta_x$, $\zeta_y$ denote the coordinates of the robot, and $\theta$ is the heading direction. In the form of~\eqref{eq:dyn}, the priori known nominal model $f$ is given by:
\begin{align}
    \zeta_x^{(t+\Delta t)} &= \zeta_x^{(t)} + \Delta t\ v\ \text{cos}(\theta^{(t)}) \notag \\
    \zeta_y^{(t+\Delta t)} &= \zeta_y^{(t)} + \Delta t\ v\ \text{sin}(\theta^{(t)}) \label{eq:dubin_car}\\
    \theta^{(t+\Delta t)} &= \theta^{(t)} + \Delta t\ u^{(t)} \notag
\end{align}
where the velocity $v=3$ and discrete time step size $\Delta t=0.1$. We also consider an unknown model-error $g$, which is bounded by $[0, 0.1]$ in $x$ and $y$ dimensions. The system is controlled by NN controllers trained by our algorithms, i.e., $u^{(t)} = \text{NN}(x^{(t)}), \; \text{NN} \in \p^{K \times b} \subset \R^{1 \times 4}$ with the controller space $\p^{K \times b}$ considered to be a hyperrectangle.

\begin{figure}[!h]
    \center
    \resizebox{.48\textwidth}{!}{
    \begin{tabular}{c}
        \includegraphics[height=0.1\textwidth]{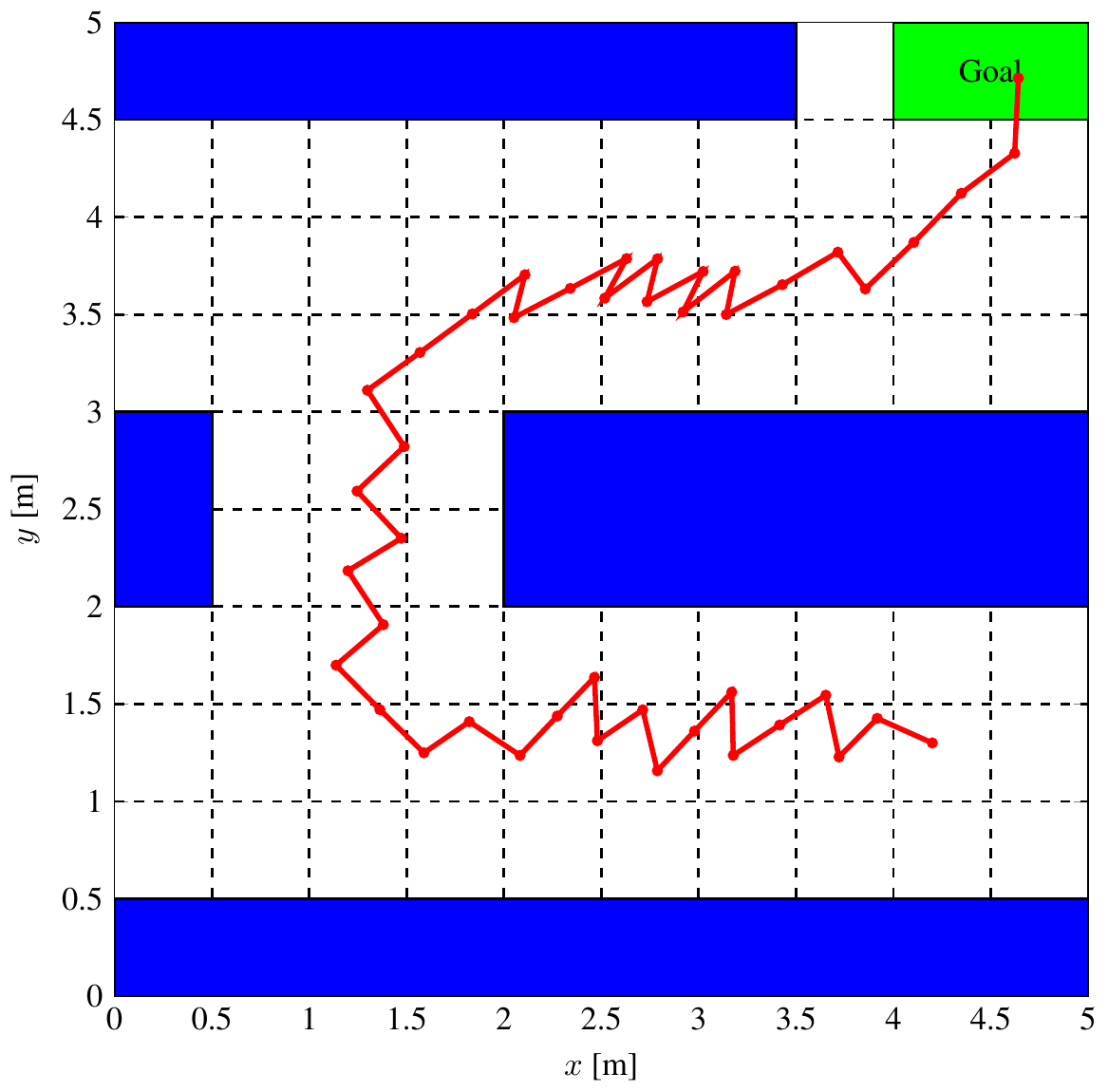}
        \includegraphics[height=0.1\textwidth]{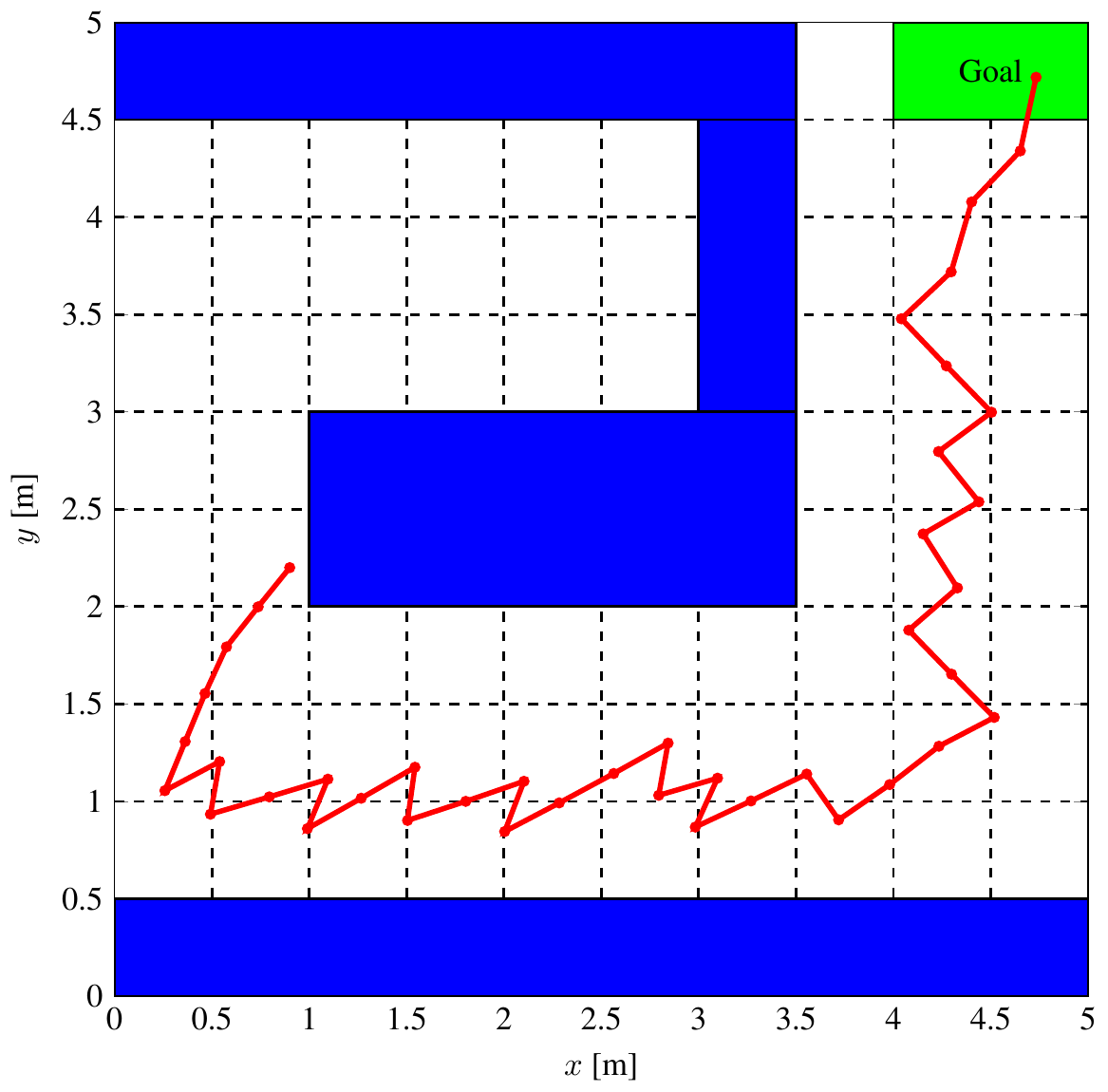}
        \includegraphics[height=0.1\textwidth]{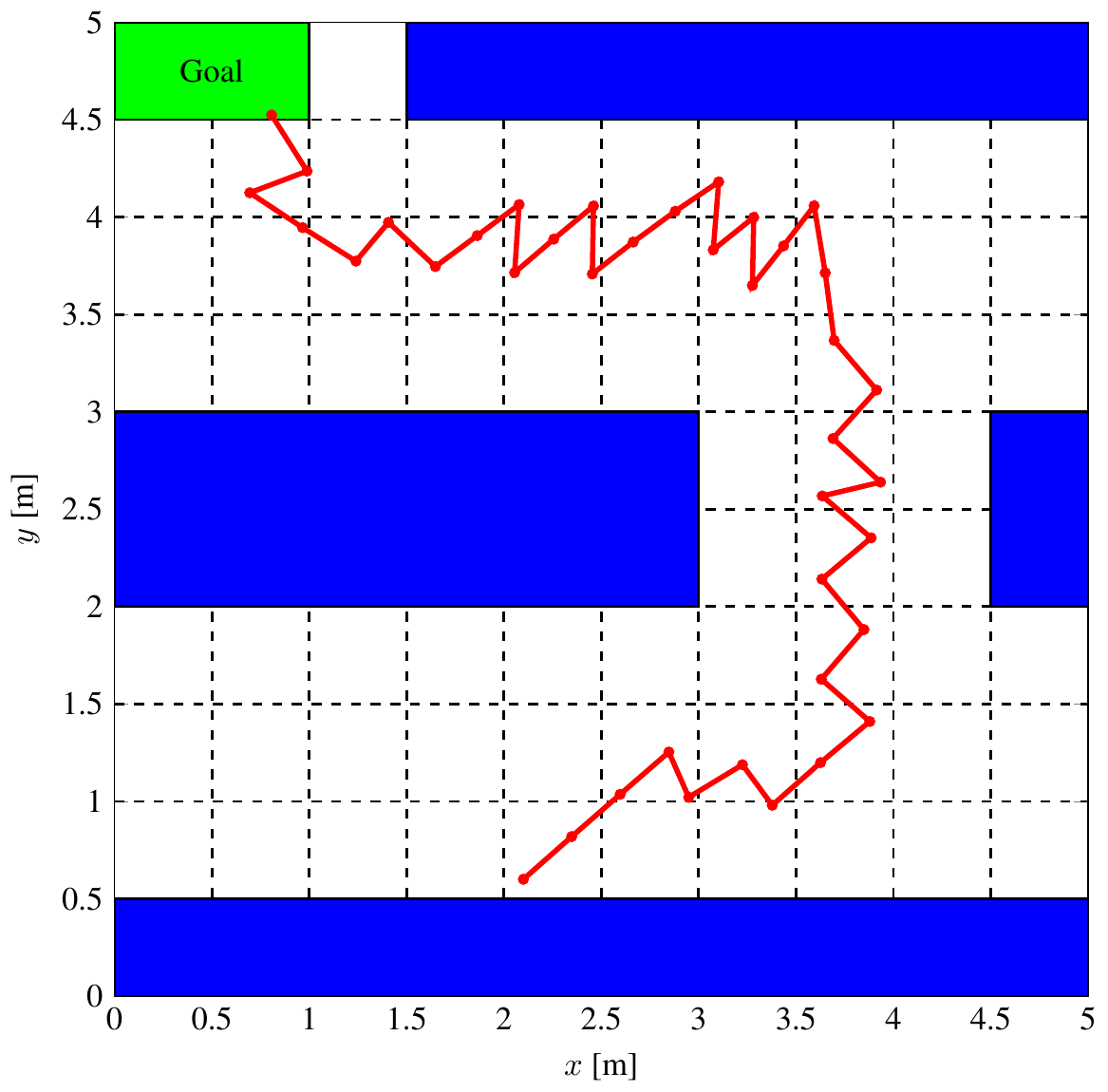} 
        \\ 
        \includegraphics[height=0.1\textwidth]{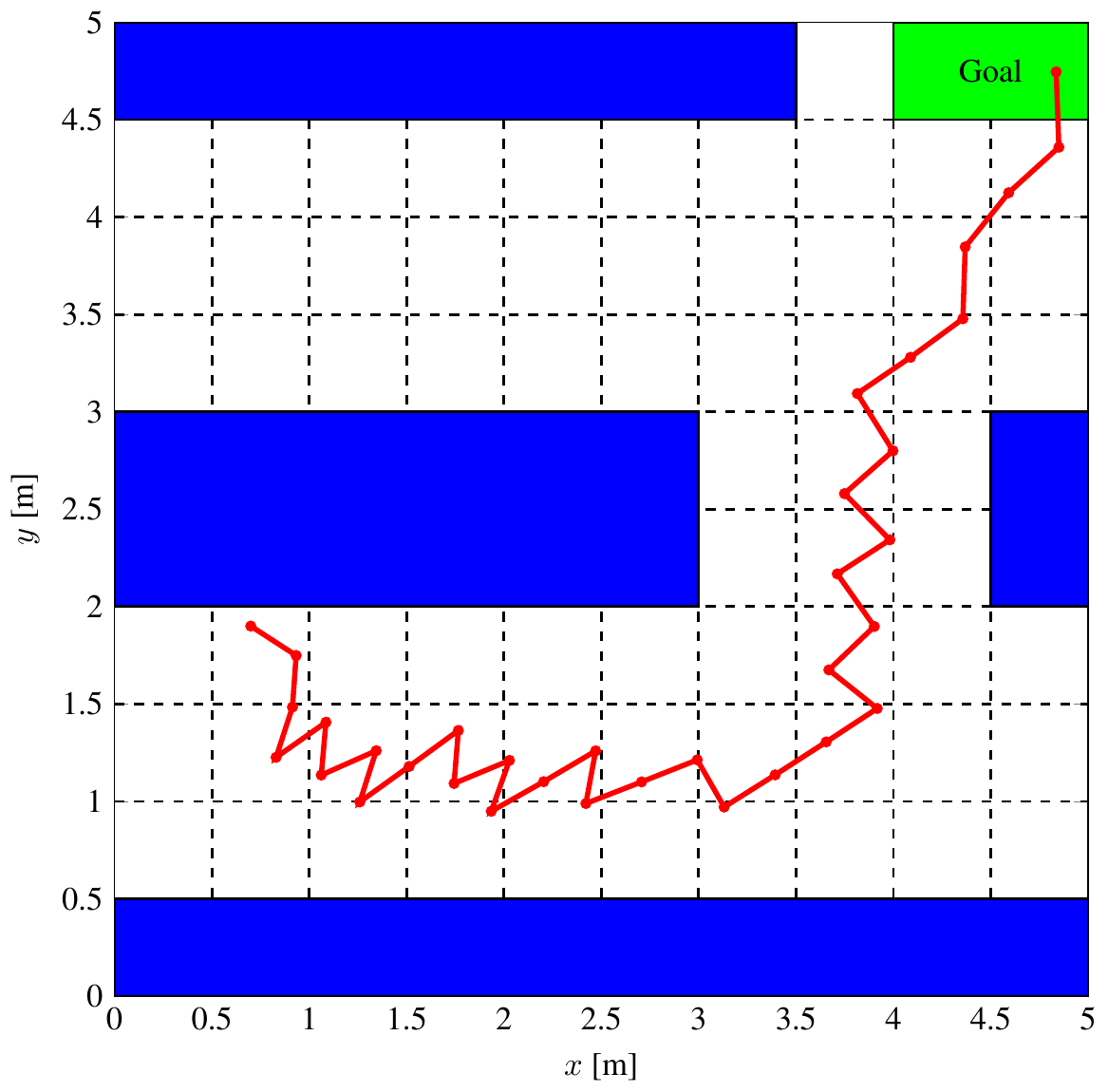}
        \includegraphics[height=0.1\textwidth]{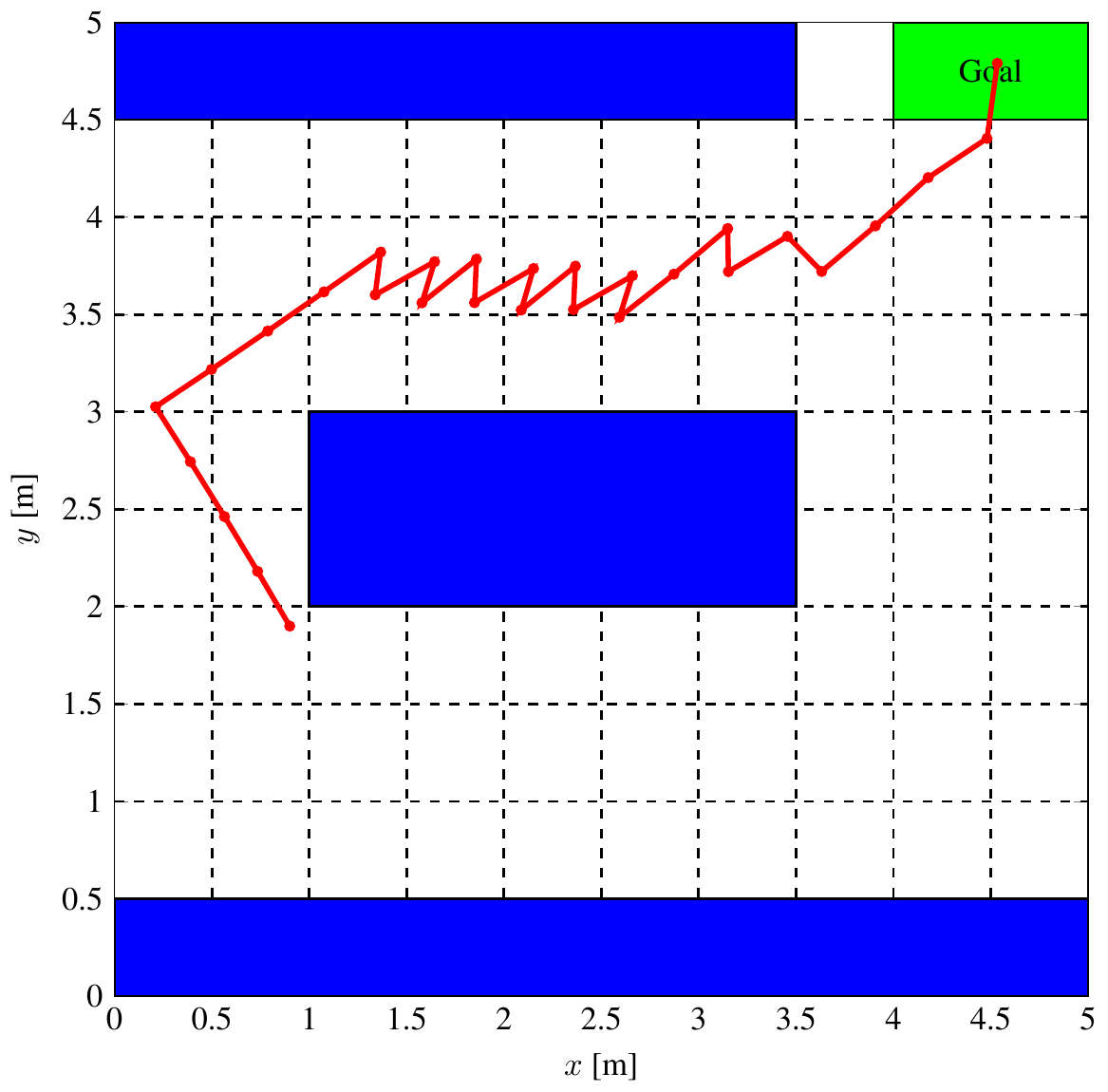}
        \includegraphics[height=0.1\textwidth]{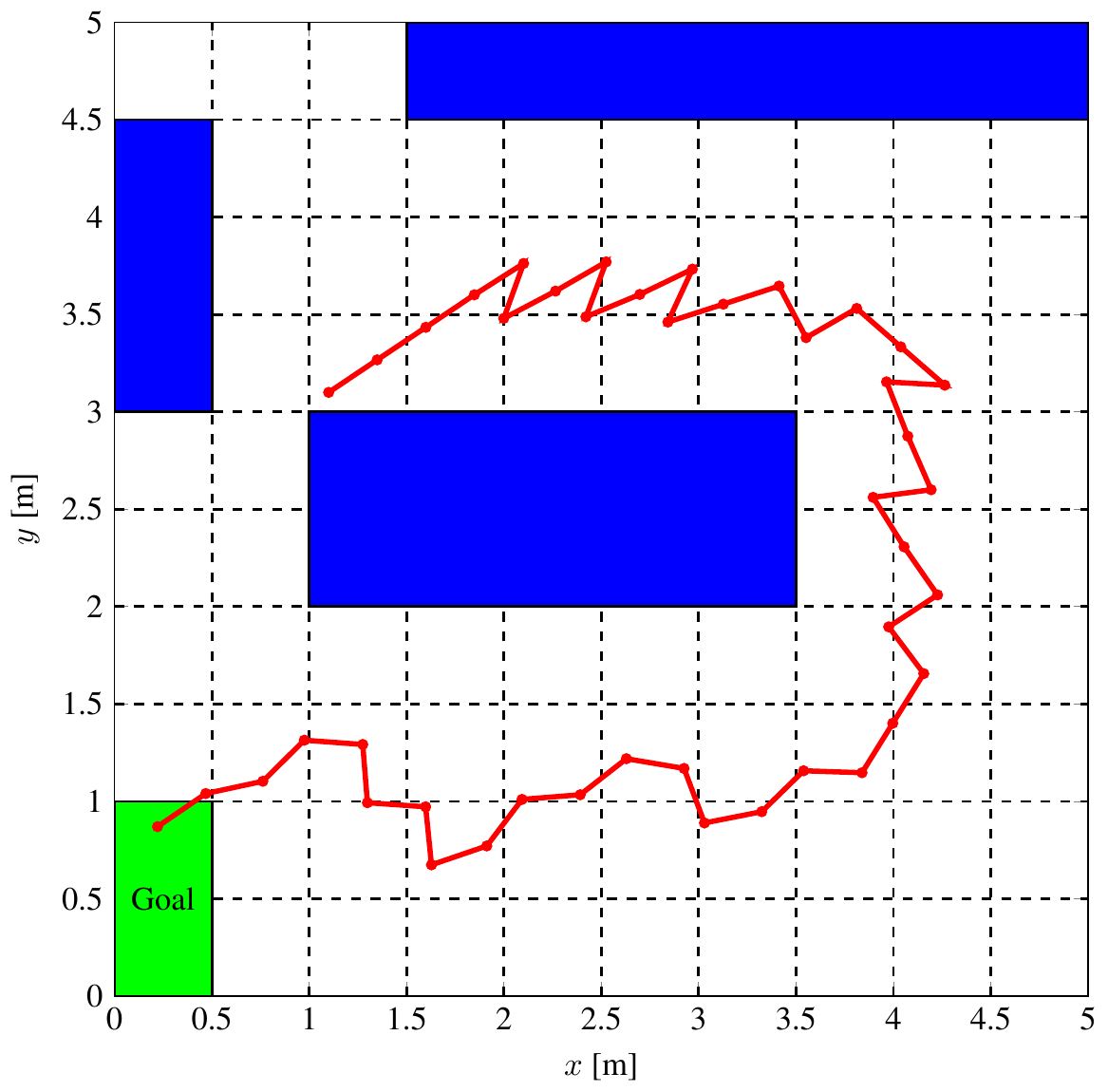}
    \end{tabular}
    }\vspace{-3mm}
    \caption{The upper row shows trajectories for Task $\W_1$, $\W_3$, $\W_5$, and the lower row corresponds to Task $\W_2$, $\W_4$, $\W_6$. The subset of local networks $\mathfrak{NN}_\text{part}$ is trained for Task $\W_1$ and the rest five tasks are given at runtime. Trajectories in all the tasks satisfy both the safety specification $\phi_\text{safety}^{\W_i}$ (blue areas are obstacles) and the liveness specification $\phi_\text{liveness}^{\W_i}$ for reaching the goal (green area).}\vspace{-4mm}
    \label{fig:traj}  
\end{figure}

As the first step of our algorithm, we discretized the state space $\mathcal{X} \subset \R^3$ and the controller space $\p^{K \times b} \subset \R^{1 \times 4}$ as described in Section~\ref{subsec:abstraction}. Specifically, we partitioned the range of heading direction $\theta \in [0, 2\pi)$ uniformly into $8$ intervals, and the partitions in the $x$, $y$ dimensions are shown as the dashed lines in Figure~\ref{fig:traj}. We uniformly partitioned the controller space $\p^{K \times b}$ into $240$ hyperrectangles. To construct the abstract model, we use GP regression to learn the model-error $g$ and use TIRA~\cite{TIRA} to compute the reachable sets.

During the offline training, we trained a subset of local networks $\mathfrak{NN}_\text{part}$ by following Algorithm~\ref{alg:online-train} in Section~\ref{sec:transfer}. Specifically, the local NNs are trained for Task $\W_1$ (the first subfigure in the upper row of Figure~\ref{fig:traj}), and the set $\mathfrak{NN}_\text{part}$ consists of $658$ local NNs. We used Keras~\cite{chollet2015keras} to train each local network, which is a shallow NN (one hidden layer) with $6$ hidden layer neurons. With PPO, each local NN is trained for $800$ episodes, and we projected the NN weights at the end of training. The total time for training and projecting weights of the $658$ local networks in $\mathfrak{NN}_\text{part}$ is $2368$ seconds.

At runtime, we tested the trained NN controller in five unseen tasks $\W_i$, $i=2, \ldots, 6$, and the corresponding trajectories are shown in Figure~\ref{fig:traj}. For each of the tasks, our framework first computes the set of safe abstract states $\X_\safe^{\W_i}$ and the function $P_\safe^{\W_i}$, and then assigns one transition at each abstract state $q \in \X_\safe^{\W_i}$. The local NNs corresponding to the assigned transitions may not have been trained offline. If this was the case, we follow Algorithm~\ref{alg:online-run} that employs transfer learning to fast learn the missing NNs at runtime. Specifically, after initializing a missing NN using the closest NN to it, we trained it for $80$ episodes, which is much less than the number of episodes used in the offline training. In Table~\ref{tab:time}, we report the execution to assign controller partitions, as well as to generate the trajectories shown in Figure~\ref{fig:traj}. For example, for Task $\W_2$, the length of the corresponding trajectory (the first subfigure in the lower row of Figure~\ref{fig:traj}) is $35$, and $28$ local NNs used along the trajectory are not in the set $\mathfrak{NN}_\text{part}$. Our algorithm efficiently trains these $28$ local NNs in $10.5$ seconds, which shows the capability of our NN controller in real-time applications.

\begin{table}[!h]
    \caption{Runtime Execution Time} \vspace{-4mm}
    \begin{center} 
    \resizebox{.99\columnwidth}{!}{
    \begin{tabular}{|c|c|c|c|c|c|c|c|c|}
    \hline
    \textbf{Task} & \textbf{Assign Controller} & \textbf{Generate Trajectory} &\textbf{Trajectory} &\textbf{\# of NNs Trained}\\ 
    \textbf{Index} & \textbf{Partitions [s]} & \textbf{Time [s]} &\textbf{Length} &\textbf{at Runtime}\\ 
    \hline\hline
    $\W_1$  & 7.1   & 0.5   & 44    & 0 \\ \cline{1-5}
    $\W_2$  & 7.6   & 10.5  & 35    & 28 \\ \cline{1-5} 
    $\W_3$  & 7.8   & 13.6  & 42    & 38 \\ \cline{1-5}
    $\W_4$  & 7.4   & 2.7   & 29    & 4  \\ \cline{1-5}
    $\W_5$  & 7.6   & 14.4  & 40    & 40 \\ \cline{1-5}
    $\W_6$  & 9.3   & 15.6  & 44    & 44 \\ \cline{1-5}
\end{tabular}       
}
\end{center}
\label{tab:time} 
\vspace{-10mm}
\end{table}



 
\bibliographystyle{IEEEtran}
\bibliography{yasser.bib, biblio.bib}    



\section*{APPENDIX}    
In this appendix, we present the proof of Theorem~\ref{thm:optimality}.

\begin{proof}
    Consider an arbitrary state $x \in q_i$ for any $q_i \in \X_\safe$, and let $z = \ct_\X(q_i)$. Recall that $\hat{V}^*_k$ is the optimal value function of the finite-state MDP $\hat{\Sigma}$ and satisfy~\eqref{eq:abst_Q}-\eqref{eq:abst_V}. Notice that
    \begin{align}
        &|V^\NN_k(x) - V^*_k(x)| \notag \\
        &\leq |V^\NN_k(x) - \hat{V}^*_k(q_i)| + |\hat{V}^*_k(q_i) - V^*_k(x)|. \label{eq:two_parts}
    \end{align}

    We first show by induction that for any $k = 0, \ldots, H$:
    \begin{equation}
    |V^\NN_k(x) - \hat{V}^*_k(q_i)| \leq (H-k) \Delta^\NN. \label{eq:Delta_nn}
    \end{equation}    
    The base case $k = H$ trivially holds since $V^\NN_H = \mathbf{1}_{\mathcal{X}_\goal}$ and $\hat{V}_H^* =  \mathbf{1}_{\X_\text{goal}}$. For the induction hypothesis, suppose for $k+1 < H$, it holds that
    \begin{equation}
        |V^\NN_{k+1}(x) - \hat{V}^*_{k+1}(q_i)| \leq (H-k-1)\Delta^\NN. \label{eq:hypothesis}
    \end{equation}
    Let $\bar{V}^*_k$ be a piecewise constant interpolation of $\hat{V}^*_k$ defined by $\bar{V}^*_k(x) = \hat{V}^*_k(q)$ for any $x \in q$ and any $q \in \X$. Then,
    \begin{align}
        &|V^\NN_k(x) - \hat{V}^*_k(q_i)|  \notag \\
        &\leq |V^\NN_k(x) - V^\NN_k(z)| + |V^\NN_k(z) - \bar{V}^*_k(z)|. \label{eq:induction}
    \end{align}
    For the first term:
    \begin{align}
        &|V^\NN_k(x) - V^\NN_k(z)| \notag \\
        &= |\mathbf{1}_{\mathcal{X}_\goal}(x) + \mathbf{1}_{\mathcal{X}_\safe^\prime}(x) \int_{\mathcal{X}_\safe} V_{k+1}^\NN (x^\prime) T(dx^\prime|x, \NN(x)) \notag \\
        &\ - \mathbf{1}_{\mathcal{X}_\goal}(z) + \mathbf{1}_{\mathcal{X}_\safe^\prime}(z) \int_{\mathcal{X}_\safe} V_{k+1}^\NN (x^\prime) T(dx^\prime|z, \NN(z))| \notag \\
        &\leq \int_{\mathcal{X}_\safe} V_{k+1}^\NN (x^\prime) |T(dx^\prime|x, \NN(x)) - T(dx^\prime|z, \NN(z)| \notag \\
        &\leq \int_{\mathcal{X}_\safe} (|t(x^\prime|x, \NN(x)) - t(x^\prime|z, \NN(x))| \notag \\
        &\qquad \qquad + |t(x^\prime|z, \NN(x)) - t(x^\prime|z, \NN(z))|) \mu(dx^\prime) \notag \\
        &\leq \Lambda_i \norm{x - z} + \Gamma_i \norm{\NN(x) - \NN(z)} \notag \\
        &\leq \Lambda_i \delta_q + \Gamma_i L_i \delta_q \label{eq:1st}
    \end{align}
    where the first inequality holds since $x, z$ are in the same abstract state $q_i$, so either $x, z \in \mathcal{X}_\goal$ or $x, z \not \in \mathcal{X}_\goal$.
    For the second term:
    \begin{align}
        &|V^\NN_k(z) - \bar{V}^*_k(z)| \notag \\
        &= |\mathbf{1}_{\mathcal{X}_\goal}(z) + \mathbf{1}_{\mathcal{X}_\safe^\prime}(z) \int_{\mathcal{X}_\safe} V_{k+1}^\NN (x^\prime) T(dx^\prime|z, \NN(z)) \notag \\
        &\ - \mathbf{1}_{\X_\text{goal}}(q) + \mathbf{1}_{\X_\safe^\prime}(q) \max_{\p \in P_\safe(q_i)} \sum_{q^\prime \in \X_\safe} \hat{V}_{k+1}^* (q^\prime) \hat{T}(q^\prime | q, \p)| \notag \\
        &\leq |\int_{\mathcal{X}_\safe} V_{k+1}^\NN (x^\prime) T(dx^\prime|z, \NN(z)) \notag \\
        &\quad - \max_{\p \in P_\safe(q_i)} \int_{\mathcal{X}_\safe} \bar{V}_{k+1}^* (x^\prime) T(dx^\prime | z, \ct_\mathbb{P}(\p)(z))| \notag \\
        &\leq \int_{\mathcal{X}_\safe} |V_{k+1}^\NN (x^\prime) - \bar{V}_{k+1}^* (x^\prime)| T(dx^\prime|z, \NN(z)) \notag \\
        &+ \max_{\p \in P_\safe(q_i)} \int_{\mathcal{X}_\safe} \bar{V}_{k+1}^* (x^\prime) |T(dx^\prime|z, \NN(z)) \notag \\
        &\hspace{41mm} - T(dx^\prime | z, \ct_\mathbb{P}(\p)(z))| \notag \\
        &\leq (H-k-1) \Delta^\NN + \sqrt{m(n+1)} \mathcal{L}_\mathcal{X} \Gamma_i \delta_\p  \label{eq:2nd}
    \end{align}
    where the last inequality uses the induction hypothesis~\eqref{eq:hypothesis} and the fact that the NN controller is aligned with the selected controller partition due to the weight projection. We also used the inequality 
    $\norm{K(x) - K^\prime(x)} \leq \norm{K - K^\prime} \norm{x} \leq \sqrt{m(n+1)} \norm{K - K^\prime}_{\max} \mathcal{L}_\mathcal{X} \leq \sqrt{m(n+1)} \delta_\p \mathcal{L}_\mathcal{X}$ for $K, K^\prime$ in the same controller partition $\p \subset \R^{m \times (n+1)}$ and $x \in \mathcal{X}_\safe$. Substitute~\eqref{eq:1st} and~\eqref{eq:2nd} into~\eqref{eq:induction} yields~\eqref{eq:Delta_nn}.

    Similarly, it can be shown that $|\hat{V}^*_k(q_i) - V^*_k(x)| \leq (H-k) \Delta^*$. Substitute this along with~\eqref{eq:Delta_nn} into~\eqref{eq:two_parts} completes the proof.
\end{proof}

\end{document}